\documentclass[sigconf]{acmart}

\usepackage{balance} 
\usepackage{hyperref}
\usepackage{url}
\usepackage{amsthm}
\usepackage{graphicx}
\usepackage{subcaption}
\usepackage{algorithm}
\usepackage{algpseudocode}
\usepackage[most]{tcolorbox}

\makeatletter
\gdef\@copyrightpermission{
  \begin{minipage}{0.2\columnwidth}
   \href{https://creativecommons.org/licenses/by/4.0/}{\includegraphics[width=0.90\textwidth]{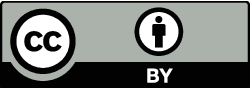}}
  \end{minipage}\hfill
  \begin{minipage}{0.8\columnwidth}
   \href{https://creativecommons.org/licenses/by/4.0/}{This work is licensed under a Creative Commons Attribution International 4.0 License.}
  \end{minipage}
  \vspace{5pt}
}
\makeatother

\title[AAMAS-2026 Formatting Instructions]{STO-RL: Offline RL under Sparse Rewards via LLM-Guided Subgoal Temporal Order}

\author{Chengyang Gu}
\affiliation{
  \institution{Hong Kong University of Science and Technology (Guangzhou)}
  \city{Hong Kong}
  \country{China}}
\email{cgu893@connect.hkust-gz.edu.cn}

\author{Yuxin Pan}
\affiliation{
  \institution{Hong Kong University of Science and Technology}
  \city{Hong Kong}
  \country{China}}
\email{yuxin.pan@connect.ust.hk}

\author{Hui Xiong}
\affiliation{
  \institution{Hong Kong University of Science and Technology (Guangzhou)}
  \city{Hong Kong}
  \country{China}}
\email{xionghui@hkust-gz.edu.cn}

\author{Yize Chen}
\affiliation{
  \institution{University of Alberta}
  \city{Edmonton}
  \country{Canada}}
\email{yize.chen@ualberta.ca}

\begin{abstract}

Offline reinforcement learning (RL) enables policy learning from pre-collected datasets, avoiding costly and risky online interactions, but it often struggles with long-horizon tasks involving sparse rewards. Existing goal-conditioned and hierarchical offline RL methods decompose such tasks and generate intermediate rewards to mitigate limitations of traditional offline RL, but usually overlook temporal dependencies among subgoals and rely on imprecise reward shaping, leading to suboptimal policies. To address these issues, we propose STO-RL (Offline RL using LLM-Guided Subgoal Temporal Order), an offline RL framework that leverages large language models (LLMs) to generate temporally ordered subgoal sequences and corresponding state-to-subgoal-stage mappings. Using this temporal structure, STO-RL applies potential-based reward shaping to transform sparse terminal rewards into dense, temporally consistent signals, promoting subgoal progress while avoiding suboptimal solutions. The resulting augmented dataset with shaped rewards enables efficient offline training of high-performing policies. Evaluations on four discrete and continuous sparse-reward benchmarks demonstrate that STO-RL consistently outperforms state-of-the-art offline goal-conditioned and hierarchical RL baselines, achieving faster convergence, higher success rates, and shorter trajectories. Ablation studies further confirm STO-RL’s robustness to imperfect or noisy LLM-generated subgoal sequences, demonstrating that LLM-guided subgoal temporal structures combined with theoretically grounded reward shaping provide a practical and scalable solution for long-horizon offline RL.

\end{abstract}


\newcommand{\BibTeX}{\rm B\kern-.05em{\sc i\kern-.025em b}\kern-.08em\TeX}

\begin{document}


\pagestyle{fancy}
\fancyhead{}


\maketitle 


\section{Introduction}
In recent years, Deep Reinforcement Learning (DRL) has emerged as a powerful paradigm for solving sequential decision-making problems, achieving remarkable success in domains ranging from game-playing \citep{goldwaser2020deep} to robotics \citep{nguyen2019review}. Despite these achievements, DRL methods usually require extensive online interaction with the environment, which can be costly, risky, or infeasible in real-world applications such as autonomous driving or medical treatment planning \citep{isele2018safe, yauney2018reinforcement}. To overcome this limitation, offline reinforcement learning (offline RL) \citep{kumar2020conservative} has been investigated, where agents learn policies entirely from pre-collected datasets without further interactions with targeting environments.

However, to date offline RL faces significant challenges in long-horizon, goal-reaching tasks with sparse rewards \citep{wu2024offline}, which are practical when agents interacting with the real world. In these settings, rewards are often provided only at the final state, and in limited offline datasets, the lack of intermediate feedback  also hinders efficient reward propagation. These  make RL agents very challenging to learn effective behaviors \citep{holmes2025attention, yunis2024subwords}. To mitigate this issue, goal-conditioned offline RL extends traditional offline RL by conditioning policies or value functions on goal representations, enabling more flexible and generalizable goal-directed behaviors \citep{ma2022offline, park2024ogbench, ding2019goal}. Complementarily, hierarchical reinforcement learning (HRL) has been proposed to tackle sparse-reward and  long-horizon problems by introducing a multi-level decision-making hierarchy \citep{schwab2017offline, nachum2018data, nachum2019does, osa2019hierarchical}. In such a framework, a high-level policy generates subgoals to guide long-term planning, while a low-level policy learns to achieve these subgoals using primitive actions. By decomposing complex tasks into sequences of subgoal-conditioned subtasks,  HRL can improve learning efficiency and exploration, but at an additional cost of algorithm design and complexity~\citep{zhang2021hierarchical}.

\begin{figure*}[htbp]
    \centering
    \includegraphics[width=0.95\linewidth]{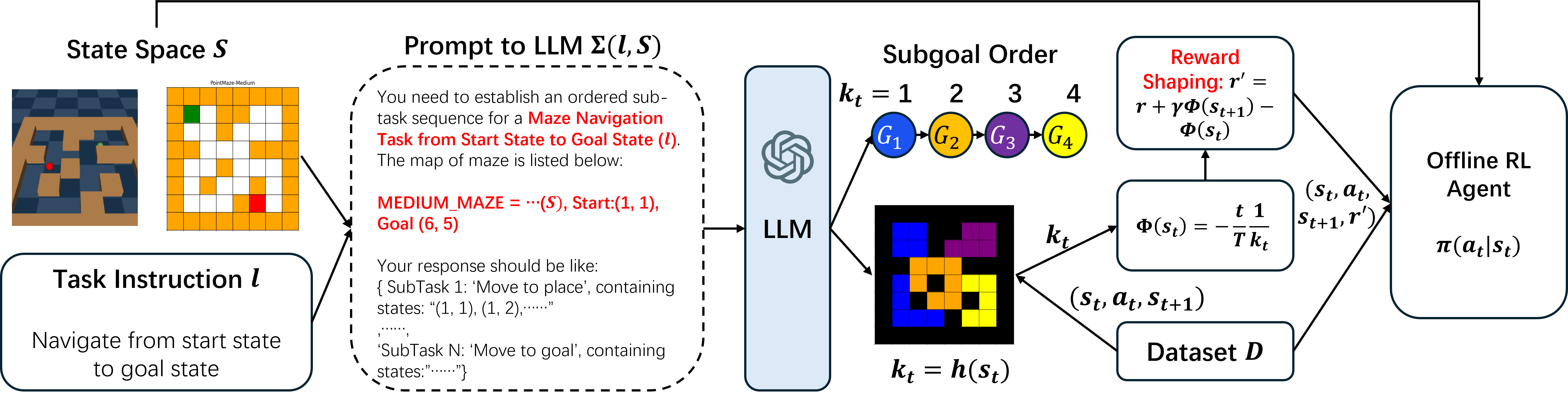} 
    \caption{Overview of STO-RL. The agent first prompts an LLM to decompose task instructions into temporally ordered subgoals $\mathcal{G}=\{G_1, G_2, ..., G_K\}$ with a mapping function $h$ linking each state $s_t$ to its subgoal index $k_t$; then applies potential-based reward shaping to assign dense rewards to each transition $(s_t, a_t, s_{t+1})$ in dataset $D$ based on the mapping function $h$, followed by training an offline RL agent on the augmented dataset to learn the optimal policy.}
    \label{fig:Overview}
\end{figure*}


Despite these advances, both offline goal-conditioned RL and offline hierarchical RL continue to face two fundamental challenges: (1) \textbf{Learning Effective Subgoals} \citep{mesbah2024subgoal, kim2021landmark, chane2021goal}. Generating informative and achievable subgoals is critical for their success, because high-level and goal-conditioned policies rely on them to guide low-level actions toward long-horizon goals, directly influencing learning efficiency and task success. However, discovering such subgoals remains difficult due to limited data coverage, sparse rewards, and the presence of suboptimal trajectories and noisy transitions in offline datasets \citep{shin2023guide, park2023hiql, wang2023state, kujanpaa2023hierarchical, rosete2023latent}. Furthermore, existing approaches often fail to model the \textbf{temporal dependencies among subgoals} \citep{baek2025graph, xu2024learning, ahn2025option}, making it challenging to learn temporally coherent subgoal sequences. This limitation leads high-level policies to generate subgoals that are inconsistent with task progression and hinder effective hierarchical coordination. (2) \textbf{Reward Shaping} \citep{mezghani2023learning, gao2024hierarchical, okudo2021subgoal}. To alleviate the challenge of sparse rewards and enhance learning efficiency, reward shaping is commonly employed to provide denser feedback signals that guide agents toward meaningful, intermediate progress. While shaping schemes such as distance-based potential functions can introduce unintended local optima and misguide policy learning, ultimately leading to suboptimal performance \citep{liu2022goal}.

In this work, we identify large language model (LLM) as the critical aid, and LLM-based planning can redefine how subgoals and rewards are discovered in long-horizon, sparse-reward problems. By leveraging their powerful natural language understanding and reasoning capabilities, LLMs can decompose complex tasks into structured, temporally ordered subgoal sequences, enabling zero-shot or few-shot plan generation \citep{yao2023react, saha2024system, shukla2023lgts}. Interestingly, recent works demonstrate that LLMs can also be applied to design effective and dense reward functions for goal-reaching tasks \citep{xie2023text2reward, pouplin2024synergy, bhambri2024extracting}. However, few studies jointly address both perspectives:  LLM-based high-level planners mostly rely on pre-trained low-level agents to execute subgoals, while LLM-based reward-shaping mechanisms also often neglect the temporal dependencies among subgoals, leading to inconsistencies and reduced reliability in hierarchical decision-making.

Motivated by these promises and gaps between LLM planners and reward shaping, in this paper we propose \textit{Offline \textbf{R}einforcement \textbf{L}earning using LLM-Guided \textbf{S}ub-goal \textbf{T}emporal \textbf{O}rder (STO-RL)}, a unified framework designed to simultaneously achieve subgoal discovery and reward shaping in the offline setting. Specifically, STO-RL prompts an LLM with task instructions and environment context to generate a sequence of temporally ordered subgoals, along with a mapping function that assigns each state to its corresponding subgoal stage. LLM excels at interpreting and generalizing these goal-reaching abstractions and relations. Leveraging this temporal structure, STO-RL applies a potential-based reward shaping scheme that integrates subgoal order into the reward function, effectively transforming sparse terminal rewards into dense signals that promotes subgoal progress to the final goal while theoretically avoiding suboptimal shaping. The resulting augmented dataset with shaped rewards enables efficient offline training of an optimal policy from a limited offline dataset. 

To validate the effectiveness of STO-RL, we conduct extensive simulations across both discrete and continuous goal-reaching benchmarks, including \textit{CliffWalking, FourRoom, PointMaze-UMaze } and \textit{PointMaze-Medium.} The results demonstrate that STO-RL significantly improves both learning efficiency and final success rates compared to state-of-the-art offline goal-conditioned and hierarchical RL baselines. In discrete environments, STO-RL achieves faster convergence and more stable policy improvement, while in continuous tasks, it consistently attains higher success rates and generates shorter, more direct trajectories toward the goals. These findings confirm that STO-RL effectively leverages LLM-guided temporally ordered subgoal structures, and, when combined with theoretically grounded reward shaping, provides a practical and scalable solution to the persistent challenges of sparse-reward goal-reaching tasks. Furthermore, ablation studies indicate that STO-RL exhibits a notable robustness to the quality of LLM-generated subgoal sequences: even when the temporal ordering is partially imperfect or noisy, STO-RL maintains competitive performance, demonstrating strong resilience to variability in LLM outputs.

\section{Related Works}
\label{gen_inst}

\paragraph{Goal-conditioned RL} Goal-conditioned reinforcement learning (RL) aims to enable agents to achieve arbitrary goals by conditioning policies and value functions on goal representations. In offline RL settings, goal-conditioned imitation learning approaches \citep{ding2019goal, ma2021goal, reuss2023goal} have been investigated to better exploit suboptimal datasets. To facilitate learning in goal-reaching tasks, goal-conditioned RL often leverages self-supervised techniques, such as hindsight relabeling \citep{andrychowicz2017hindsight, li2020generalized}, vector quantization methods \citep{islam2022discrete, ozair2021vector, kujanpaa2023hierarchical} and chain-of-thought temporal abstraction via Transformers \citep{jia2024chain}, to derive informative subgoal representations. Despite these advances, existing goal-conditioned RL approaches can suffer from suboptimal or noisy subgoal selection \citep{yang2022rethinking}, as learning these goals autonomously can be nontrivial. 

\paragraph{Hierarchical RL}
Hierarchical Reinforcement Learning (HRL) has achieved notable success in addressing long-horizon goal-reaching problems under sparse reward settings by decomposing complex tasks into subgoals. A key factor in its effectiveness is the learning of appropriate subgoal representations, which significantly influences both performance and stability \citep{liu2021hierarchical}. Recent studies have explored different strategies for discovering semantically meaningful subgoals efficiently. For example, Guider and HLPS \citep{wang2024probabilistic, shin2023guide} leverage pre-trained probabilistic distributions to predict subgoals, while HIDIO \citep{zhang2021hierarchical} and SEADS \citep{achterhold2023learning} adopt self-supervised approaches based on intrinsic rewards. Another line of research, including HIGL \citep{kim2021landmark}, SFL \citep{hoang2021successor}, constructs task graphs from landmarks to guide subgoal selection \citep{huang2019mapping}. Although these approaches demonstrate promising progress in subgoal discovery and skill learning, they largely overlook the temporal structure among subgoals, which constrains their efficiency and overall effectiveness. This limitation motivates our use of external knowledge sources, such as Large Language Models (LLMs), to uncover temporal structures among subgoals and guide hierarchical policy learning.

\paragraph{LLM Planning} Large Language Models (LLMs) have recently emerged as powerful tools for high-level planning in sequential decision-making tasks. By leveraging their pretrained world knowledge \citep{yildirim2024task, singh2025adaptbot} and step-by-step reasoning capabilities \citep{guo2025deepseek, ferrag2025llm}, LLMs can capture the causal and temporal dependencies inherent in goal-directed behavior. This enables them to generate structured plans composed of subgoals with explicit temporal order \citep{song2023llm, logeswaran2022few, yao2023react, liu2023llm+}, providing effective intermediate guidance for hierarchical reinforcement learning agents. For example, ELLM \citep{du2023guiding} rewards agents for achieving LLM-suggested goals based on their current state, while LDSC \citep{shek2025option} and LgTs \citep{shukla2023lgts} train specialized RL agents to achieve subgoals generated by language models. Despite these advances, LLMs are limited to high-level planning and cannot directly execute actions, requiring low-level agents to achieve each subgoal along the specified temporal sequence \citep{zhou2025step, prakash2023llm, li2025llm}. One intrinsic issues for LLM-based planners are that they lack direct interations with low-level policies. If the pre-trained fine-grained policies are not well-aligned with the LLM-generated high-level plan, agents may fail to follow the intended subgoal progression. To address this, we introduce a temporal-order-aware reward shaping framework that integrates LLM-generated subgoal sequences into offline RL, allowing agents to autonomously learn subgoal completion in the correct temporal order toward the final goal.

\paragraph{Reward Shaping} Reward shaping augments the ground-truth reward with an additional shaping term, typically in an additive form. Several studies employ manually designed shaping rewards to accelerate policy learning by incorporating domain expertise~\citep{colombetti1992robot, randlov1998learning}. However, such methods require considerable problem-specific design, and human knowledge fails to consistently yield gains during policy training. In contrast, numerous methods adopt exploration bonuses or intrinsic rewards to expedite policy convergence without relying on domain knowledge~\citep{bellemare2016unifying, ostrovski2017count, zheng2018learning, memarian2021self, devidze2022exploration, ma2025catching}. However, these approaches often cause the learned policy to deviate from the original optimal policy. To guarantee policy invariance, potential-based reward shaping (PBRS) is formulated as the temporal difference of a potential function~\citep{ng1999policy}. Building on this foundation, subsequent work has focused on the theoretical analysis of PBRS~\citep{wiewiora2003potential, laud2003influence} and proposed variants that offer more informative guidance. These include action-dependent PBRS~\citep{wiewiora2003principled}, dynamic PBRS~\citep{devlin2012dynamic}, dynamic action-dependent PBRS~\citep{harutyunyan2015expressing}, mode-based PBRS~\citep{asmuth2008potential}, dynamics-aware PBRS~\citep{koprulu2025dense}, and self-shaped rewards~\citep{adamczyk2025bootstrapped}. Compared to prior PBRS algorithms, our method investigates PBRS within the hierarchical RL setting, where the shaping reward incorporates domain knowledge extracted from a LLM.

\section{Method}
\label{Method}

\subsection{Problem Formulation}
In this work, we focus on the offline reinforcement learning (RL) for goal-reaching tasks with sparse rewards. It can be formulated as a finite-horizon Markov Decision Process (MDP): $M = (\mathcal{S}, \mathcal{A}, P, r, \gamma, T)$. We also set up the problem with a collaborative language-based task description $l$ and a pre-collected dataset $\mathcal{D}$. Here, $\mathcal{S}, \mathcal{A}$ denotes the state space and the action space, respectively; $P(s_{t+1}| s_t,a_t)$ represents the transition dynamics; $r:\mathcal{S}\times\mathcal{A}\times\mathcal{S}\to\mathbb{R}$ is the reward function; $\gamma \in (0,1)$ is the discount factor; and $T$ is the episode horizon length. The reward function $r$ is given by
\begin{align}
    r(s_t, a_t, s_{t+1}) = \left\{ \begin{array}{ll} 
1 & \text{when } s_t = g, \\ 
0 & \text{otherwise}.
\end{array} \right.
\end{align}
with $g \in \mathcal{S}$ denoting the target goal state.

The dataset $\mathcal{D}$ consists of a collection of trajectories: \[\tau=\{(s_0, a_0), (s_1, a_1), ..., (s_{T-1}, a_{T-1}), s_T\},\]
And our goal is to learn an optimal policy agent $\pi(s_t|a_t): \mathcal{S} \to \mathcal{A}$, that maximizes the expected discounted return:
\[V^\pi = \mathbb{E}_\pi[ \sum_{t=0}^{T-1} \gamma^t r(s_t,a_t,s_{t+1})].\]

\subsection{LLMs-Guided Temporal Ordered Subgoals}
Standard offline RL methods often struggle to perform effectively on goal-reaching tasks due to sparse rewards and long planning horizons, which hinder the identification of meaningful gradients during policy optimization. A natural approach to mitigate these challenges is to decompose the task into a sequence of subgoals that are easier to achieve, allowing the agent to receive denser feedback and learn structured representations of progress. Such idea of hierarchical decomposition has long been central to reinforcement learning, inspiring the development of subgoal discovery and hierarchical policy learning frameworks.

Due to its strong reasoning and understanding capabilities, LLMs have been increasingly applied to high-level planning in robotics and embodied decision-making \citep{chen2025robogpt, li2024embodied, wu2023embodied}. When prompted with natural language task instructions together with necessary state and environment descriptions, LLMs can decompose tasks into structured subgoal sequences. Owing to their strong common sense reasoning and causal interpretations, these LLM-generated plans inherently capture the \textbf{temporal dependencies} among subgoals. However, prior research often overlook such temporal structure, which limits their effectiveness in RL tasks.

In this paper, we assume that for any goal-reaching task, there exists an ordered subgoal sequence of $K$ \textbf{temporally ordered subgoals}: $\{G_1, G_2, \dots, G_K\} \subseteq \mathcal{G}$, such that every successful trajectory reaching the final goal $g$ must achieve these subgoals sequentially. The formal definition of such a trajectory is stated in \textit{Definition 1}.

\paragraph{Definition 1.} \textit{(Properties of Successful Trajectory)\; Consider a  trajectory $\tau = \{(s_0, a_0,k_0), ..., (s_{H-1}, a_{H-1},k_{H-1})\}$  successfully reaching the final goal $g$ in $H$ timesteps ($H \leq T$). $\tau$ should satisfy: (1) $k_0 = 1, k_{H-1} = K$; (2) There exists $K-1$ timesteps $0 < t^1 < ...,< t^{K-1} < H-1$, where $k_{t^i} = i, k_{t^i+1} = i+1$}.

We define the \emph{progress index} as $k_t \in \{1,\dots,K\}$, where $k_t$ represents the temporal order of the corresponding subgoal at timestep $t$. Given the language-based task instruction $l$ and state space $\mathcal{S}$ for environment description, we construct a language prompt $\Sigma(l, \mathcal{S})$ that is fed into a large language model (LLM). Prior studies \citep{liu2023llm+, prakash2023llm, silver2022pddl} demonstrate that LLMs can reliably generate subgoals from such prompts. An example of prompt is illustrated in Figure~\ref{fig:Overview}. Unlike prior works, we let LLM output a subgoal sequence $\mathcal{G}$ along with an additional mapping function $h: \mathcal{S} \to \{1, 2, ...,K\}$ which associates each state $s_t$ with its corresponding progress index $k_t=h(s_t)$. This mapping provides a foundation for incorporating subgoal temporal order into the reward structure and policy learning process, enabling more efficient and interpretable offline RL in complex, sparse-reward environments.

\subsection{Subgoal-Temporal-Order-Aware Potential-based Reward Shaping }
\label{sec:rewardshape}
Built upon the temporally ordered subgoal sequences generated by LLMs, we propose a novel reward shaping mechanism that transforms the original sparse reward signal into a dense one, effectively accelerating learning while avoiding suboptimal solutions in goal-reaching tasks.

\paragraph{Definition 2.} \textit{(Positive Progress) Let $\{G_1, G_2, \dots, G_K\}$ denote the sequence of subgoals arranged in prescribed temporal order. Define $k_t \in \{1, 2, \dots, K\}$ as the corresponding progress index at time step $t$. A transition $(s_t, a_t, s_{t+1})$ is said to constitute\textbf{ positive progress} if and only if $k_{t} < k_{t+1}$.}

The key insight of our method is to exploit the \emph{temporal structure} inherent in the ordered subgoal sequence, closely aligned with LLM's strong capabilities. As established in \textit{Definition 1}, any trajectory that successfully reaches the final goal must pass through these subgoals sequentially. Consequently, an optimal policy should reach each subgoal as early as possible within the episode. To formalize this intuition, we first consider the reward design at the transition level. Specifically, following the insights of previous works \citep{qiu2023instructing, baek2025graph}, transitions that \textbf{promote progress toward the next subgoal} are considered more desirable. We formally bring out a definition of such transitions as \textit{positive progress} transitions in \textit{Definition 2}.

We introduce a \emph{potential-based reward shaping} (PBRS) function \citep{ng1999policy} that encodes both temporal urgency and ordered subgoal progress: $\Phi(s_t) = -\frac{t}{T} \cdot \frac{1}{k_t}$, where $T$ denotes the maximum episode length. The shaped reward $r'$ is then defined as:
\begin{equation}
    r'(s_t,a_t,s_{t+1}) = r(s_t,a_t,s_{t+1}) + \gamma \Phi(s_{t+1}) - \Phi(s_t). 
    \label{eq:reward}
\end{equation}

By introducing this potential-based shaping, we not only preserve the optimality of the original policy, but also effectively identify and reward actions that guide the agent more efficiently toward the goal. As is shown in \textit{Theorem 1}, transitions that achieve positive subgoal progress receive strictly higher shaped rewards, and they are therefore preferred during learning, enabling more efficient policy optimization.

\paragraph{Theorem 1. } \textit{(Shaped Reward Preference for Positive Progress) Consider the shaped reward in Eq. \ref{eq:reward}: $r'(s_t,a_t,s_{t+1}) = r(s_t,a_t,s_{t+1}) + \gamma \Phi(s_{t+1}) - \Phi(s_t)$ with potential function $\Phi(s_t) =  -\frac{t}{T} \cdot \frac{1}{k_t}$. Then, for any transition $(s_t,a_t,s_{t+1})$, the shaped reward assigns strictly higher value to transitions that achieve positive progress compared to those that do not, namely $\Delta r:= r'(s_t, a_t^c, s_{t+1}^c) - r'(s_t, a_t^n, s_{t+1}^n)>0$.}
\begin{proof}
Let $(s_t, a_t^c, s_{t+1}^c)$ denote a transition that achieves \textbf{positive progress}, and $(s_t, a_t^n, s_{t+1}^n)$ a transition that does \textbf{not}. 

Substituting $\Phi(s) = -\tfrac{t}{T}\cdot \tfrac{1}{k_t}$ to the definition of reward shaping, we have:
\begin{align}
    r'(s_t, a_t^c, s_{t+1}^c)  &= r(s_t, a_t^c, s_{t+1}^c) - \gamma (\frac{t+1}{T} \cdot \frac{1}{k_{t+1}^c}) + \frac{t}{T} \cdot \frac{1}{k_t}, \\
    r'(s_t, a_t^n, s_{t+1}^n)  &= r(s_t, a_t^n, s_{t+1}^n) - \gamma (\frac{t+1}{T} \cdot \frac{1}{k_{t+1}^n}) + \frac{t}{T} \cdot \frac{1}{k_t}. 
\end{align}

Taking the difference we have:
\begin{equation}
\begin{split}
        \Delta r &= r'(s_t, a_t^c, s_{t+1}^c) - r'(s_t, a_t^n, s_{t+1}^n) \\
    &=  r(s_t, a_t^c, s_{t+1}^c) - r(s_t, a_t^n, s_{t+1}^n) + \gamma \frac{t+1}{T} \cdot (\frac{1}{k_{t+1}^n} - \frac{1}{k_{t+1}^c}).
\end{split}
\end{equation}

Since the task has sparse rewards, indicating
\begin{align}
    r(s_t, a_t^c, s_{t+1}^c) = r(s_t, a_t^n, s_{t+1}^n) = 0.
\end{align}

By the definition of positive progress in \textit{Definition 1}:
\begin{equation}
    k_{t+1}^n \leq k_{t} < k_{t+1}^c.
\end{equation}

Therefore:
\begin{equation}
    \Delta r= \gamma \tfrac{t+1}{T}\Bigl(\tfrac{1}{k_{t+1}^n} - \tfrac{1}{k_{t+1}^c}\Bigr) > 0 .
\end{equation}
showing that shaped rewards are strictly larger for transitions achieving positive progress.  
\end{proof}

In addition to promoting faster learning, our temporally ordered reward shaping also enhances the agent’s ability to avoid suboptimal solutions. Specifically, it provides a mechanism to differentiate between successful trajectories of varying lengths, favoring shorter ones that reach the goal more efficiently.

\paragraph{Lemma 1.} \textit{(Equivalence of Returns for Equal-Length Successful Trajectories) Consider a goal-reaching task with $K$ temporally ordered subgoals. Any two successful trajectories with same episode length $T_L$ ($T_L\leq T$) yield equal discounted return $\sum_{t=0}^{T_L-1}\gamma^tr'(s_t,a_t,s_{t+1})$ under $r'(s_t,a_t,s_{t+1}) = r(s_t,a_t,s_{t+1}) + \gamma \Phi(s_{t+1}) - \Phi(s_t)$, where $\Phi(s_t) = -\frac{t}{T} \cdot \frac{1}{k_t}$.}

\begin{proof}
See Appendix 6.1.
\end{proof}

\begin{figure}[t]
    \centering
    \includegraphics[width=1\linewidth]{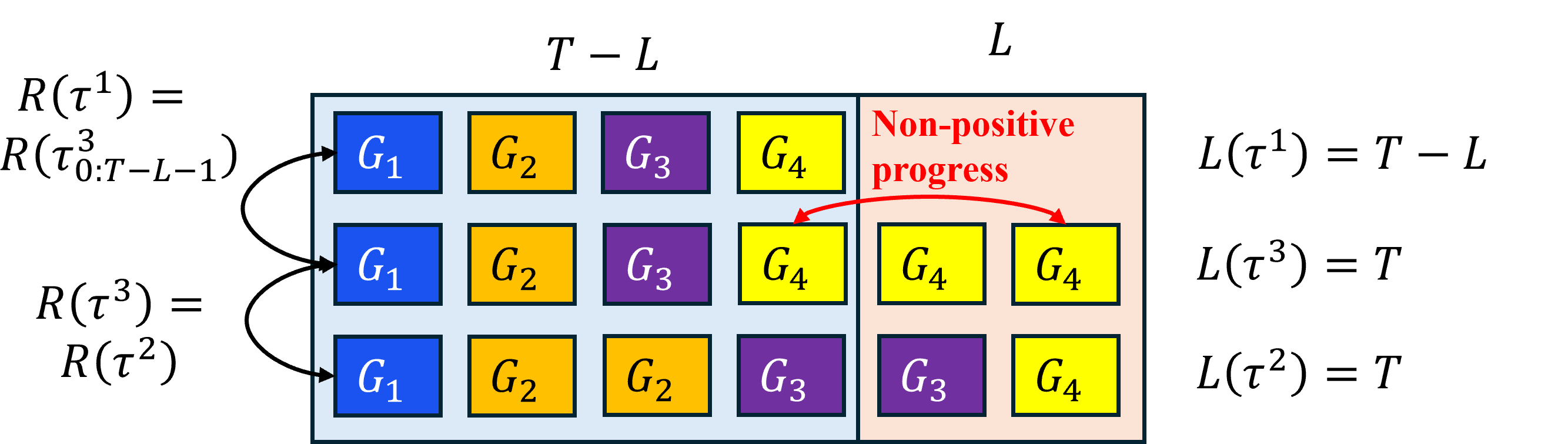}
    \caption{A toy example illustrating how the subgoal–temporal–order–aware PBRS penalizes longer successful trajectories. A virtual trajectory $\tau^3$ is constructed with the same length $T$ and discounted return as longer trajectory $\tau^2$ ($R(\tau^3)=R(\tau^2)$), consisting of the first $T-L$ steps sharing subgoal indices and return with shorter $\tau^1$, followed by $L$ non-progress transitions that yield negative returns under \textit{Theorem 2}.}
    \label{fig:Toy}
\end{figure}

Based on such potential-based reward shaping properties as shown in \textit{Lemma 1}, any two successful trajectories with the same length will yield the same discounted return under the shaped reward, regardless of the specific sequence of intermediate subgoal progress indices. Furthermore, under the condition (discount factor $\gamma$ satisfies $\gamma \geq (T-1)/T$), \textit{Theorem 2} demonstrates that transitions which do not contribute to positive subgoal progress receive strictly negative rewards. Consequently, for two successful trajectories of lengths $T$ and $T-L$ ($L>0$), the discounted return of the longer trajectory can be expressed as the sum of the shorter trajectory’s return and the cumulative rewards of $L$ non-progressive transitions, as illustrated in Figure~\ref{fig:Toy}. Since these additional transitions yield negative contributions, the return of the longer trajectory is strictly smaller than that of the shorter one, as proven in \textit{Theorem 3}.

\paragraph{Theorem 2.} \textit{(Strict Negative Reward for Non-Positive Progress): If $\gamma > \frac{T-1}{T}$, potential-based penalty $\gamma \Phi(s_{t+1}) - \Phi(s_t) < 0$ holds for any transition $(s_t, a_t, s_{t+1})$ that does not achieve positive progress: $k_{t+1} \leq k_t$.}
\begin{proof}
Consider the potential function defined as $\Phi(s_t) = - \frac{t}{T} \cdot \frac{1}{k_t}$, where $t \in \{0, \dots, T-1\}$ is the timestep and $k_t$ is the progress index.  

For a transition $(s_t, a_t, s_{t+1})$ that does \textbf{not achieve positive progress}, we have $k_{t+1} \le k_t$.

The potential-based shaping term is

\begin{equation}
\begin{split}
\Delta \Phi_t 
&:= \gamma \Phi(s_{t+1}) - \Phi(s_t) \\
&= \gamma \left(- \frac{t+1}{T} \cdot \frac{1}{k_{t+1}}\right) 
   - \left(- \frac{t}{T} \cdot \frac{1}{k_t}\right) \\
&= - \gamma \frac{t+1}{T} \cdot \frac{1}{k_{t+1}} 
   + \frac{t}{T} \cdot \frac{1}{k_t}.
\end{split}
\end{equation}

For any non-progress transition, Since $k_{t+1} \le k_t$, we have
\begin{equation}
    \frac{1}{k_{t+1}} \ge \frac{1}{k_t}.
\end{equation}

Hence
\begin{equation}
    \Delta \Phi_t \leq - \gamma \frac{t+1}{T} \cdot \frac{1}{k_t} + \frac{t}{T} \cdot \frac{1}{k_t} 
= \frac{1}{k_t} \left( \frac{t}{T} - \gamma \frac{t+1}{T} \right)
= \frac{1}{k_t} \cdot \frac{t - \gamma (t+1)}{T}.
\end{equation}

Since $t \le T-1$, we have
\begin{equation}
    \gamma > \frac{T-1}{T} > \frac{t}{t+1} .
\end{equation}

Then the numerator $t - \gamma (t+1) < 0$, which implies
\begin{equation}
    \Delta \Phi_t < 0.
\end{equation}

This proves that the potential-based reward penalizes transitions that fail to achieve positive progress when $\gamma > \frac{T-1}{T}$.
\end{proof}

This property ensures that our reward shaping framework inherently discourages redundant or inefficient behaviors, guiding the agent to prioritize shorter, optimal trajectories toward the goal.

\paragraph{Theorem 3.} \textit{(Shorter Successful Trajectories Yield Higher Returns): Let $\tau^1$ and $\tau^2$ be two successful trajectories with length $T_L-L$ and $T_L$ ($L>0$) respectively. If $\gamma > \frac{T-1}{T}$, then the discounted return of the shorter trajectory is strictly larger, i.e. $R(\tau^1) > R(\tau^2)$, where $R(\tau^1)=\sum_{t=0}^{T_L-L-1}\gamma^tr'(s^1_t,a^1_t,s^1_{t+1})$, $R(\tau^2)=\sum_{t=0}^{T_L-1}\gamma^tr'(s^2_t,a^2_t,s^2_{t+1})$}.

\begin{proof}
    Suppose there exists a successful trajectory $\tau^3$ with length $T_L$. In $\tau^3$, $k_{T_L-L}=k_{T_L} = K$, and for $t=T_L-L, .., T_L-1$, $k_{t+1}=k_t=K$. Hence, the discounted return of $\tau^3$ is:
    \begin{equation}
        \begin{split}
            R(\tau^3) &= \sum_{t=0}^{T_L-1}\gamma^tr'(s^3_t,a^3_t,s^3_{t+1}) \\
            & = \sum_{t=0}^{T_L-L-1}\gamma^tr'(s^3_t,a^3_t,s^3_{t+1}) + \sum_{t=T_L-L}^{T_L-L-1}\gamma^tr'(s^3_t,a^3_t,s^3_{t+1}).
        \end{split}
    \end{equation}
    Since $K_{T_L-L} =k_{T_L} = K$, according to \textit{Lemma 1} we have:
    \begin{align}
        R(\tau^3) &= R(\tau^2),\\
        R(\tau^1) & = \sum_{t=0}^{T_L-L-1}\gamma^tr'(s^3_t,a^3_t,s^3_{t+1}) .
    \end{align}
    Hence:
    \begin{equation}
        R(\tau^2) = R(\tau^1) + \sum_{t=T_L-L}^{T_L-L-1}\gamma^tr'(s^3_t,a^3_t,s^3_{t+1});
    \end{equation}
    Since $k_{t+1}=k_t$ for $t=T_L-L, .., T_L-1$, according to \textit{Theorem 2}:
    \begin{equation}
        r'(s_t, a_t, s_{t+1}) <0, \quad t=T_L-L, .., T_L-1.
    \end{equation}
    Therefore, $R(\tau^2)<R(\tau^1)$.
\end{proof}

\subsection{STO-RL}

Built upon the temporally ordered subgoal sequences generated by LLMs and the potential-based reward shaping introduced in Section~\ref{sec:rewardshape}, we propose \textit{Offline \textbf{R}einforcement \textbf{L}earning using LLM-Guided \textbf{S}ub-goal \textbf{T}emporal \textbf{O}rder (STO-RL)}, a framework for offline goal-conditioned reinforcement learning.

To ensure the RL agent is receiving rich and meaningful rewards, STO-RL first prompts an LLM to extract information from the language-based task instruction and environment description. LLM aids in decomposing the overall task into a temporally ordered subgoal sequence. Using this sequence, the framework applies potential-based reward shaping incorporating subgoal temporal order, to assign a reward signal $r'_t$ for every transition $(s_t,a_t, s_{t+1})$ in the offline dataset $D$. With the augmented dataset containing the dense rewards, an offline RL agent is trained to learn the optimal policy $\pi(s_t|a_t)$. This process allows the agent to efficiently leverage subgoal temporal information while performing policy optimization in a purely offline setting.

The complete STO-RL algorithm is summarized in Algorithm~\ref{alg:STO-RL}.

\begin{algorithm}[htbp]
\caption{STO-RL}
\label{alg:STO-RL}
\begin{algorithmic}[1]
\Require Offline dataset $\mathcal{D}$, task instruction $l$, state space $S$, discount factor $\gamma$, maximum episode length $T$
\Ensure Optimal policy $\pi^*(s_t|a_t)$
\State \textbf{Generate Prompt:} Design prompt $\Sigma=(l, \mathcal{S})$ for LLM.
\State \textbf{Subgoal Extraction:} LLM outputs temporally ordered subgoal sequence $G$ and mapping function $h$: $(G, h) = LLM(\Sigma)$ .
\For{each transition $(s_t, a_t, s_{t+1}) \in \mathcal{D}$}
    \State $\Phi(s_t) = - \frac{t}{T} \cdot \frac{1}{k_t}$ where $k_t = h(s_t)$
    \State $r'_t = r(s_t,a_t,s_{t+1}) + \gamma \Phi(s_{t+1}) - \Phi(s_t)$
\EndFor
\State \textbf{Augment dataset:} Replace original rewards in $\mathcal{D}$ with shaped rewards $r'_t$.
\State \textbf{Offline RL Training:} Train an offline RL agent to learn optimal policy $\pi^*(s_t|a_t)$.
\end{algorithmic}
\end{algorithm}

\section{Experiments}
\label{sec::experiments}
To thoroughly assess the effectiveness of the proposed STO-RL method, we conduct experiments across \textit{four} distinct goal-reaching tasks under sparse reward settings. Our evaluation is designed to address the following research questions:
\begin{enumerate}
    \item Does STO-RL achieve higher \textbf{success rates} on sparse-reward tasks in comparison with existing (hierarchical) offline reinforcement learning baselines?
    \item Will STO-RL accelerate the RL agent learning process, compared to prior methods?
    \item Can STO-RL help avoid suboptimal solutions, achieving shorter trajectory solutions?
\end{enumerate}

To generate ordered subgoal sequences in our STO-RL, we majorly employ \textit{ChatGPT 5.0} as the default LLM model (See Sec. \ref{sec::ablstudy} for other LLM models); For the offline RL component in our STO-RL framework, we adopt common-used \textit{Implicit Q-Learning (IQL)} \citep{kostrikov2021offline}, which learns value functions to infer an implicit policy via advantage-weighted regression. Across all experiments, we employ a two-layer multilayer perceptron (MLP) architecture with 128 hidden units per layer for the learning of value, Q, and policy networks in IQL. We find this architecture effective to learn Q value.

\paragraph{Baselines} We compare our STO-RL framework with 3 baseline approaches:
\begin{itemize}
    \item \textbf{Implicit Q-Learning (IQL)} \citep{kostrikov2021offline}: a classical offline RL algorithm without hierarchical subgoal decomposition.
    \item \textbf{Goal-Conditioned Behavioral Cloning (GC-BC)} \citep{ding2019goal}: a behavioral cloning style approach that learns a direct mapping from states, subgoals to actions via supervised imitation. In our experiments, GC-BC utilizes the same LLM-generated subgoals as STO-RL for a fair comparison.
    \item \textbf{Hierarchical Implicit Q-Learning (HIQL)} \citep{park2023hiql}: a hierarchical extension of offline goal-conditioned RL that derives a representation function, high-level policy, and low-level policy jointly from a single goal-conditioned value function.
\end{itemize}

\subsection{Tasks with Finite, Discrete Action Space}

\begin{figure}[t]
    \centering
    \begin{subfigure}[t]{0.48\linewidth}
        \centering
        \includegraphics[width=\linewidth]{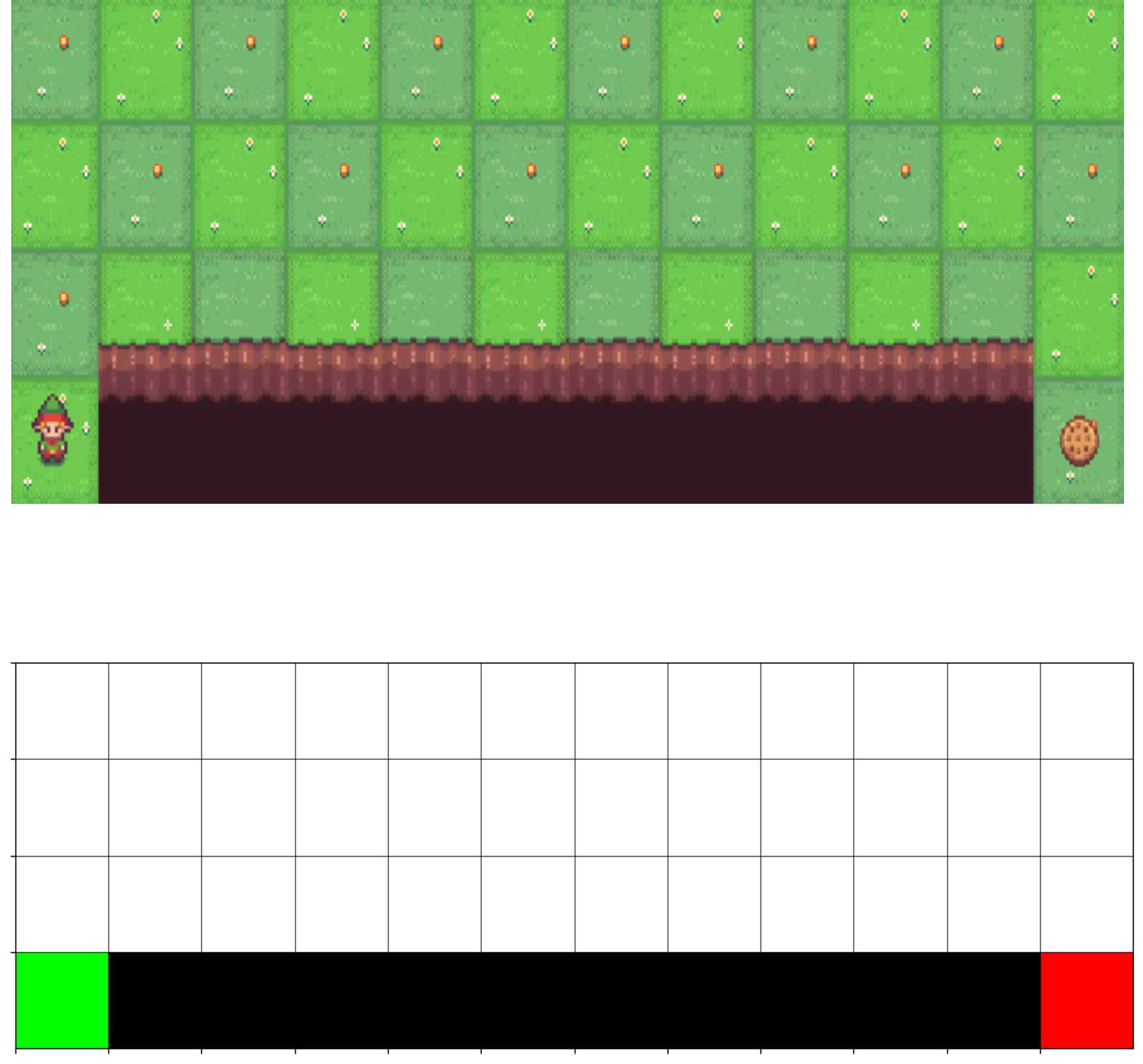}
        \caption{\textit{CliffWalking}}
        \label{fig:cliffwalking}
    \end{subfigure}
    \hfill
    \begin{subfigure}[t]{0.48\linewidth}
        \centering
        \includegraphics[width=\linewidth]{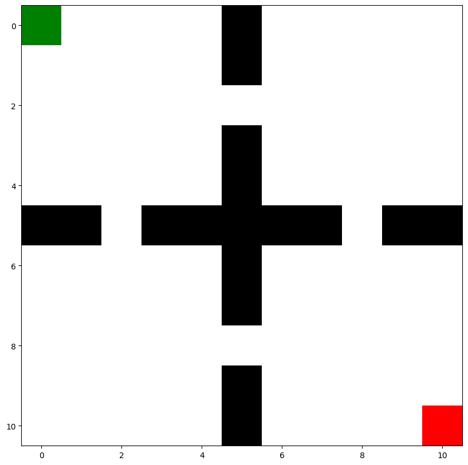}
        \caption{\textit{FourRoom}}
        \label{fig:fourroom}
    \end{subfigure}
    \caption{Environments with finite, discrete action spaces.}
    \label{fig:EnvsDiscrete}
\end{figure}

To answer the three questions, we begin by evaluating STO-RL on two simple but classical sparse-reward, goal-reaching tasks with \textbf{discrete, finite state and action spaces}: (1) \textit{CliffWalking} and (2) \textit{FourRoom}. The configurations of these environments are illustrated in Figure~\ref{fig:EnvsDiscrete}.

In the \textit{CliffWalking} task, the agent must navigate a $4\times12$ grid world from a designated start state to a goal state while avoiding falling off a cliff. The task begins with the agent positioned at $[3, 0]$ (green state in Figure~\ref{fig:cliffwalking}), and the goal is located at $[3, 11]$. A cliff occupies cells $[3,1 - 10]$ (black region in Figure~\ref{fig:cliffwalking}). If the agent steps onto a cliff cell, it is immediately returned to the start state; Similarly, in the \textit{FourRoom} task, the agent navigates an $11\times11$ grid world divided into four connected rooms with narrow doorways. The agent must move from a fixed start location in the upper-left room (green state in Figure~\ref{fig:fourroom}) to a goal location in the bottom-right room (red state in Figure~\ref{fig:fourroom}). Walls (black cells in Figure~\ref{fig:fourroom}) separate the rooms, and the agent can transition between rooms only through the designated doorways. For both tasks, the agent operates in a discrete action space consisting of four actions: {move up, move down, move right, move left}. The agent receives a reward of 1 only upon reaching the goal, and an episode terminates either when the goal is reached or when the maximum number of steps ($T=100$) is exceeded.

\begin{table}[htbp]
\small
\centering
\caption{Evaluation results on CliffWalking and FourRoom. Success rates and average steps after 1,000 training iterations.}
\label{tab:ResultsDiscrete}
\begin{tabular}{l|cc|cc}
\toprule
\textbf{Method} & \multicolumn{2}{c}{\textbf{CliffWalking}} & \multicolumn{2}{c}{\textbf{FourRoom}} \\
\cmidrule(lr){2-3} \cmidrule(lr){4-5}
 & Success Rate & Steps & Success Rate & Steps \\
\midrule
Datasets & 0.50 & 69.9 $\pm$ 32.8& 0.14 & 97.2$\pm$ 9.2\\
IQL & 1.00& 13.0 $\pm$ 0.0& 1.00& 20.0 $\pm$ 0.0\\
GC-BC &1.00 & 13.0 $\pm$ 0.0& 1.00& 20.0 $\pm$ 0.0\\
HIQL &1.00 & 13.0 $\pm$ 0.0& 1.00& 20.0 $\pm$ 0.0\\
\textbf{STO-RL} &1.00 & 13.0 $\pm$ 0.0& 1.00& 20.0 $\pm$ 0.0\\
\bottomrule
\end{tabular}
\end{table}

Following the similar settings of \cite{jiang2023offline}, we generate offline datasets for both tasks, each containing 1,000 trajectories, and evaluate STO-RL alongside the baseline methods on these datasets. Following \textit{Theorem 2}, the discount factor $\gamma$ is set to $0.99$ ($\geq(T-1)/T$) for both environments. The results in Table~\ref{tab:ResultsDiscrete}, averaged over 100 trials, demonstrate that STO-RL achieves competitive success rates compared with the baselines. STO-RL converges to the same optimal path with shortest length and 100\% success rate as all other methods, indicating no losses in baseline offline RL performance. Furthermore, the learning curves in Figure 4 show that our STO-RL always requires fewer iterations to converge to the optimal solution, indicating a \textbf{faster learning process} relative to the baselines.

\begin{figure}[htbp]
    \centering
    \begin{subfigure}[t]{0.48\linewidth}
        \centering
        \includegraphics[width=\linewidth]{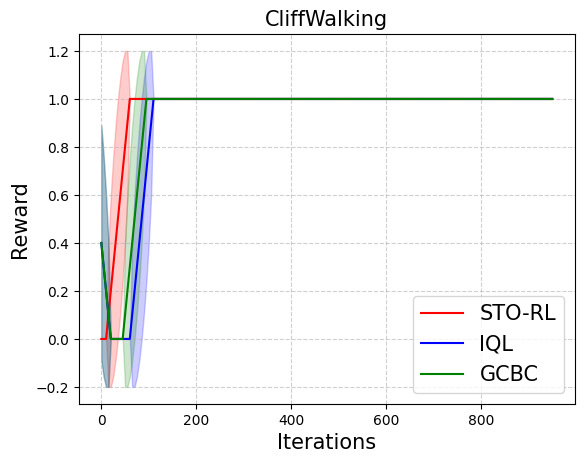}
        \caption{\textit{CliffWalking}}
        \label{fig:cliffwalkingCurve}
    \end{subfigure}
    \hfill
    \begin{subfigure}[t]{0.48\linewidth}
        \centering
        \includegraphics[width=\linewidth]{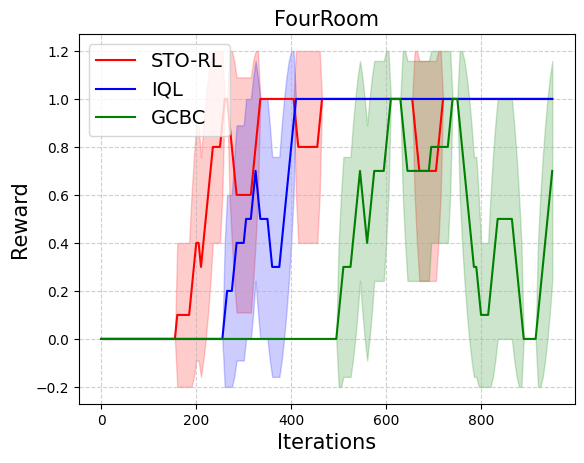}
        \caption{\textit{FourRoom}}
        \label{fig:fourroomCurve}
    \end{subfigure}
    \caption{Learning curves for the \textit{CliffWalking} and \textit{FourRoom} tasks, smoothed using a moving average over 50 iterations.}
    \label{fig:CurvesDiscrete}
\end{figure}

\begin{figure}[htbp]
    \centering
    \begin{subfigure}[t]{0.36\linewidth}
        \centering
        \includegraphics[width=\linewidth]{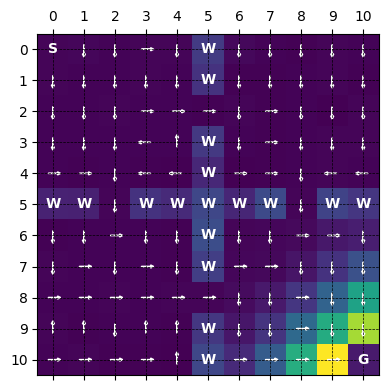}
        \caption{Value function in IQL}
        \label{fig:IQLMap}
    \end{subfigure}
    \hfill
    \begin{subfigure}[t]{0.48\linewidth}
        \centering
        \includegraphics[width=\linewidth]{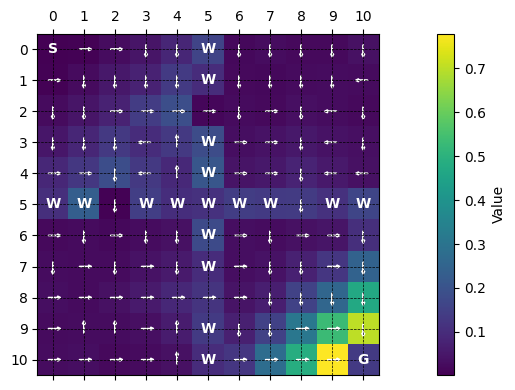}
        \caption{Value function in STO-RL}
        \label{fig:STOMap}
    \end{subfigure}
    \caption{Value functions for the \textit{FourRoom} task. ‘S’ denotes the start state, ‘G’ the goal state, and ‘W’ the walls. Arrows indicate the optimal actions under the learned policy.}
    \label{fig:ValueMap}
\end{figure}

The value maps for the \textit{FourRoom} environment, presented in Figure~\ref{fig:ValueMap}, illustrate why STO-RL achieves a faster learning process. Compared with the IQL value map in Figure~\ref{fig:IQLMap}, the subgoal temporal order based reward shaping in STO-RL produces a more accurate value landscape (Figure~\ref{fig:STOMap}), which better differentiates unrewarded states in the early stages of the task. In Figure~\ref{fig:STOMap}, the gradually increasing values along the paths to each room’s doorway guide the agent to efficiently move toward the goal, enabling rapid identification of the optimal trajectory.

\subsection{Tasks with Contiuous Action Space}
\begin{figure}[t]
    \centering
    \begin{subfigure}[t]{0.47\linewidth}
        \centering
        \includegraphics[width=\linewidth]{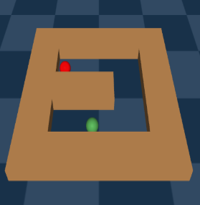}
        \caption{PointMaze-UMaze}
        \label{fig:UMaze}
    \end{subfigure}
    \hfill
    \begin{subfigure}[t]{0.48\linewidth}
        \centering
        \includegraphics[width=\linewidth]{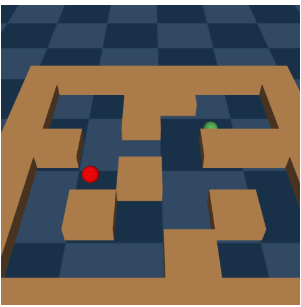}
        \caption{PointMaze-Medium}
        \label{fig:Medium}
    \end{subfigure}
    \caption{Environments with continuous state, action spaces.}
    \label{fig:EnvsContinuous}
\end{figure}

In the \textit{CliffWalking} and \textit{FourRoom} tasks, we demonstrate the faster learning capability of STO-RL as addressed in Question (2). To better evaluate answer Question (1) and (3), we assess STO-RL on two additional, more challenging sparse-reward, goal-reaching tasks from D4RL \citep{fu2020d4rl}: \textit{PointMaze-UMaze} and \textit{PointMaze-Medium}, which feature continuous state and action spaces. The configurations of these environments are illustrated in Figure~\ref{fig:EnvsContinuous}.

\begin{table}[htbp]
\centering
\small
\setlength{\tabcolsep}{4pt} 
\caption{Evaluation results on PointMaze-UMaze and PointMaze-Medium. Success rates and average steps after 2000 training iterations.}
\label{tab:ResultsContinuous}
\begin{tabular}{l|cc|cc}
\toprule
\textbf{Method} & \multicolumn{2}{c}{\textbf{UMaze}} & \multicolumn{2}{c}{\textbf{Medium}} \\
\cmidrule(lr){2-3} \cmidrule(lr){4-5}
 & Success Rate & Steps & Success Rate & Steps \\
\midrule
Datasets & 0.49 & 191.2 $\pm$ 12.7& 0.51 & 415.1$\pm$ 93.4\\
IQL      & 0.24 & 199.3 $\pm$ 0.0 & 0.17 & 467.7 $\pm$ 74.7\\
GC-BC    & 0.21 & 198.1 $\pm$ 4.4 & 0.58 & 406.8 $\pm$ 88.2\\
HIQL     & 0.47 & 197.0 $\pm$ 4.0 & 0.38 & 416.6 $\pm$ 62.4\\
\textbf{STO-RL} & 0.68 & 195.1 $\pm$ 5.3 & 0.55 & 380.7 $\pm$ 109.0\\
\bottomrule
\end{tabular}
\end{table}

In both PointMaze-UMaze and PointMaze-Medium, the agent is modeled as a 2-DoF ball actuated by forces in the Cartesian $x$ and $y$ directions. The task requires the agent to navigate a closed maze to reach a designated goal. In \textit{UMaze}, the maze has a U-shaped layout, while \textit{Medium} presents a more complex structure with multiple corridors and obstacles, increasing the difficulty of planning and exploration. To maintain consistency, the start and goal states are fixed in both environments: for \textit{UMaze}, the start is [-1, 1] with added noise and the goal is [-1, -1] with noise; for \textit{Medium} the start is [-2.5, 2.5] with noise and the goal is [1.5, -2.5] with noise, where the noise is sampled from a 2D Gaussian distribution with standard deviations (0.25, 0.25). In both tasks, the agent has a 4-dimensional state space (comprising the $x$ and $y$ positions and velocities) and and a 2-dimensional action space corresponding to the linear forces applied along $x$ and $y$ directions. A sparse reward of 1 is provided only when the agent reaches the goal (i.e., when the Euclidean distance between the agent and goal is less than 0.5). Episodes terminate either upon reaching the goal or when the maximum number of steps is exceeded ($T=200$ for \textit{UMaze}, and $T=500$ for \textit{Medium}) is exceeded, requiring agents to efficiently explore and identify optimal trajectories.  


To facilitate the generation of ordered subgoal sequences by the LLMs, the continuous maze maps in the \textit{PointMaze} tasks are discretized into cells with a side length of 1. These discretized maps are then vectorized in matrix form then provided as part of the prompt to the LLMs (See the details in Appendix 6.2).

\begin{figure}[htbp]
    \centering
    \begin{subfigure}[t]{0.48\linewidth}
        \centering
        \includegraphics[width=\linewidth]{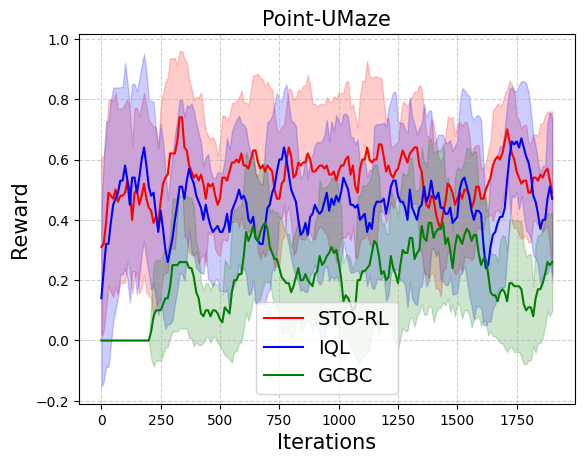}
        \caption{\textit{UMaze}}
        \label{fig:UMazeCurve}
    \end{subfigure}
    \hfill
    \begin{subfigure}[t]{0.48\linewidth}
        \centering
        \includegraphics[width=\linewidth]{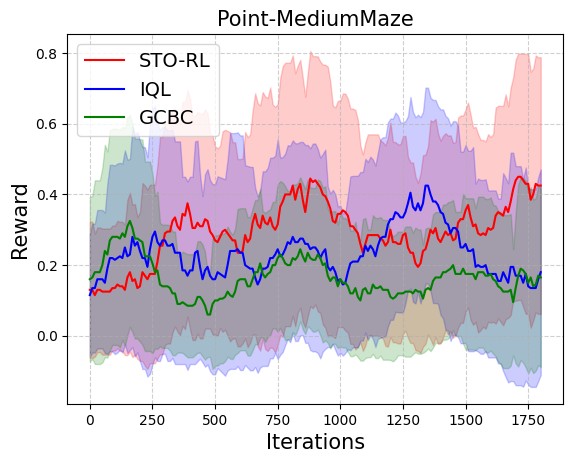}
        \caption{\textit{Medium}}
        \label{fig:MediumCurve}
    \end{subfigure}
    \caption{Learning curves for the \textit{PointMaze-UMaze} and \textit{PointMaze-Medium} tasks, smoothed using a moving average over 50 iterations.}
    \label{fig:CurvesContinuous}
\end{figure}

We generate offline datasets for both tasks, each containing 1,000 trajectories, and evaluate STO-RL alongside the baseline methods on these datasets. Following \textit{Theorem 2}, the discount factor $\gamma$ is set to $0.996$ for \textit{UMaze} and $0.999$ for \textit{Medium} ($\geq(T-1)/T$). The results in Table~\ref{tab:ResultsContinuous}, averaged over 100 trials, demonstrate that our STO-RL outperforms the baselines in both \textit{PointMaze} tasks, by achieving the most competitive performance with almost highest success rates (0.68 for \textit{UMaze} and 0.55 for \textit{Medium}) and shortest solution trajectories under given offline datasets. In \textit{Medium}, although GC-BC attains competitive success rates compared to our STO-RL, its average trajectory length is approximately 20 steps longer. Such result indicates that our STO-RL framework have better capability in avoiding suboptimal solutions. Moreover, the learning curves in Figure~\ref{fig:CurvesContinuous} indicate that our STO-RL maintains its faster learning priorities in the \textit{PointMaze} tasks. We exclude HIQL from the learning curve comparison because it trains a separate high-level policy on additional data, unlike GC-BC, IQL, and STO-RL which only train low-level agents, making direct comparison unfair. The \textit{PointMaze} tasks feature longer horizons and larger, continuous action spaces, which make exploration and learning effective actions challenging. As a result, baselines like GC-BC and IQL achieve poorer performance with limited offline data. In contrast, our STO-RL consistently attains high performance, demonstrating superior adaptability to complex offline RL tasks. Additionally, we show that STO-RL also consistently achieves superior performance across different LLM-generated subgoal sequences, as discussed in detail in Section~\ref{sec::ablstudy}.

\subsection{Ablation Study: Subgoal Sequences}
\label{sec::ablstudy}
An important factor that could significantly affect proposed STO-RL's performance  is the \textbf{quality} of the subgoal sequences generated by the LLMs. Different types of language models may produce varying subgoal sequences, potentially influencing task performance. A key question, therefore, is \textit{whether the language model can consistently generate effective subgoal sequences with correct temporal ordering, particularly in complex tasks involving environmental interference.}

\begin{figure}[htbp]
    \centering
    \includegraphics[width=0.85\linewidth]{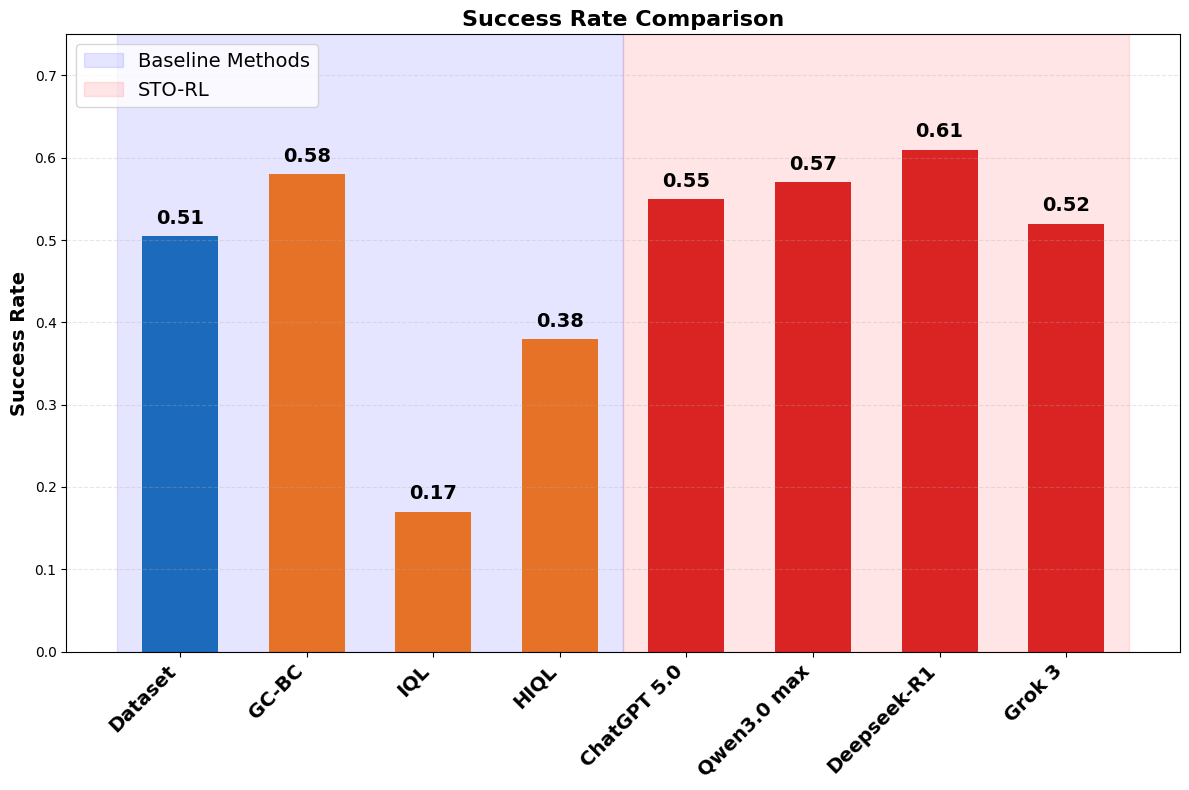}
    \caption{Success rates of STO-RL under subgoal sequences generated by different LLMs, for \textit{PointMaze-Medium}. }
    \label{fig:STOLLM}
\end{figure}

\begin{figure}[htbp]
    \centering
    \begin{subfigure}[t]{0.32\linewidth}
        \centering
        \includegraphics[width=\linewidth]{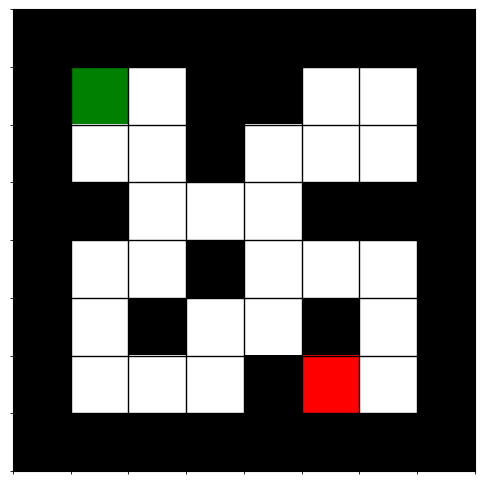}
        \caption{\textit{Medium} Maze.}
        \label{fig:MediumGrid}
    \end{subfigure}
    \hfill
    \begin{subfigure}[t]{0.32\linewidth}
        \centering
        \includegraphics[width=\linewidth]{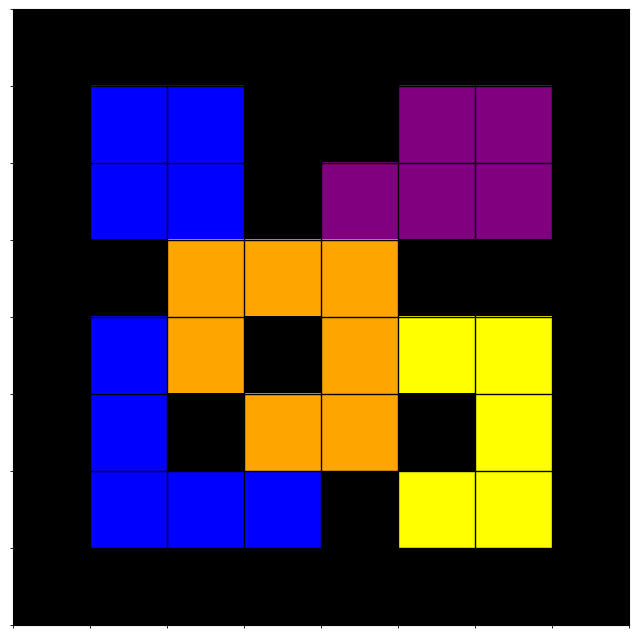}
        \caption{ChatGPT 5.0}
        \label{fig:ChatGPT}
    \end{subfigure}
    \hfill
    \begin{subfigure}[t]{0.32\linewidth}
        \centering
        \includegraphics[width=\linewidth]{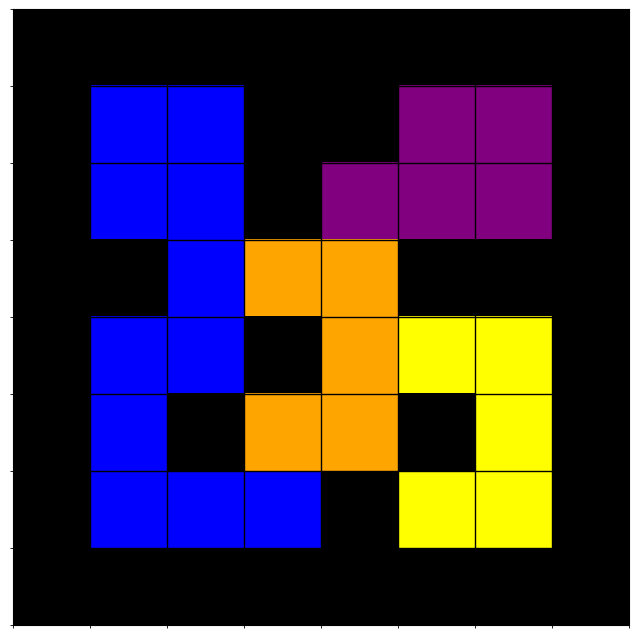}
        \caption{Grok 3}
        \label{fig:Grok}
    \end{subfigure}
    \hfill
    \begin{subfigure}[t]{0.32\linewidth}
        \centering
        \includegraphics[width=\linewidth]{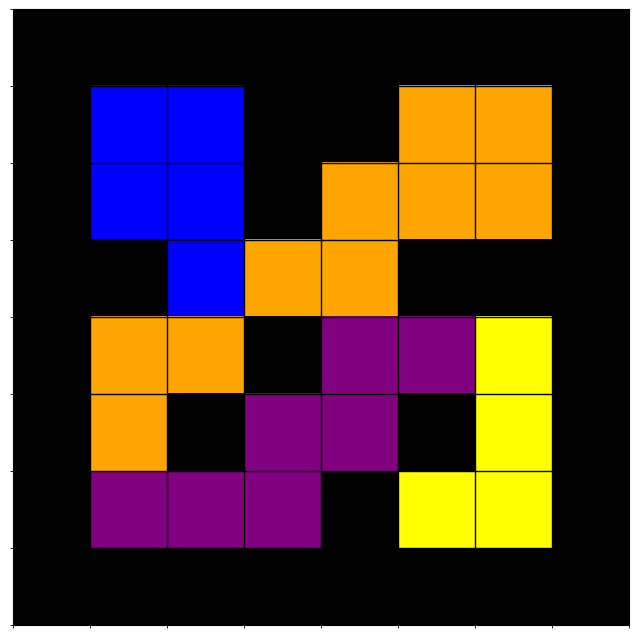}
        \caption{DeepSeek R1}
        \label{fig:DeepSeek}
    \end{subfigure}
    \hfill
    \begin{subfigure}[t]{0.32\linewidth}
        \centering
        \includegraphics[width=\linewidth]{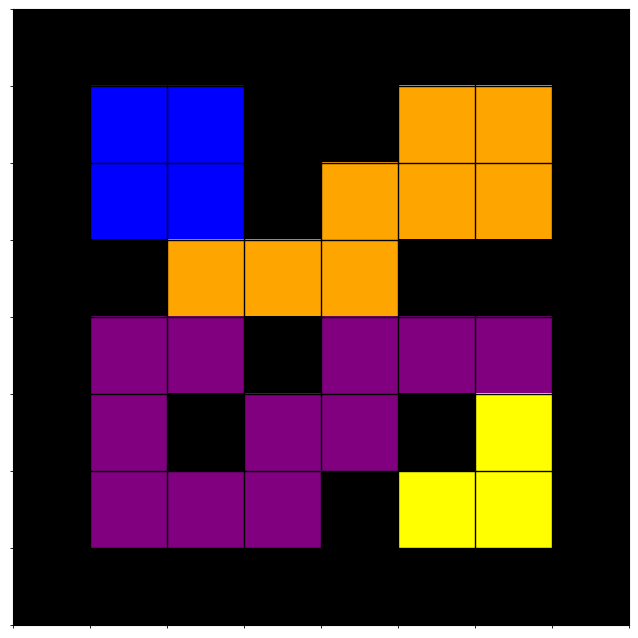}
        \caption{Qwen3-max}
        \label{fig:Qwen}
    \end{subfigure}
    \hfill
    \begin{subfigure}[t]{0.32\linewidth}
        \centering
        \includegraphics[width=\linewidth]{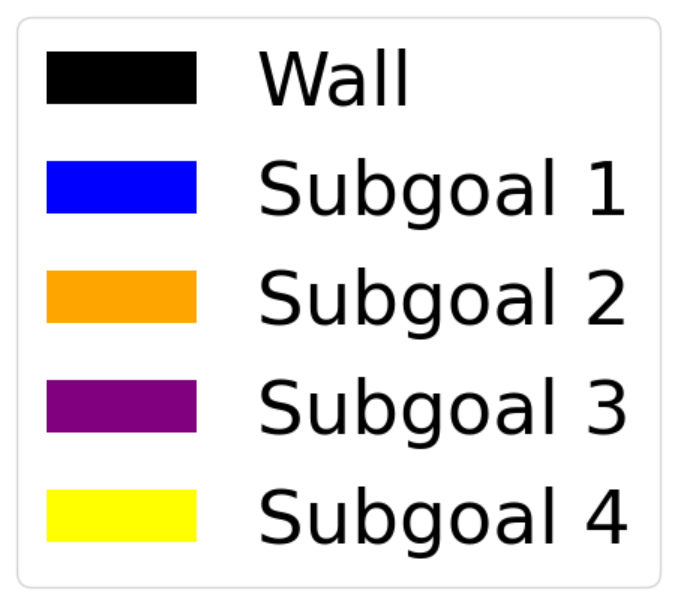}
        \caption{Subgoal Order}
        \label{fig:legend}
    \end{subfigure}
    \caption{Subgoal Sequences generated by different LLMs. In discretized \textit{PointMaze-Medium} (Figure~\ref{fig:MediumGrid}), green state is start, red state is goal. }
    \label{fig:SubGoals}
\end{figure}

To evaluate the general effectiveness of LLM-generated subgoals, we conduct an ablation study on \textit{PointMaze-Medium} task. In this study, four widely used language models—\textit{ChatGPT 5.0} \citep{openai2025gpt5systemcard}, \textit{Grok 3} \citep{xai2025grok3}, \textit{DeepSeek-R1} \citep{guo2025deepseek}, and \textit{Qwen3-Max} \citep{yang2025qwen3}—are prompted to generate distinct subgoal sequences using same instructions, which are then individually applied within our STO-RL framework. Figure~\ref{fig:SubGoals} presents the subgoal states on the maze map corresponding to the sequences generated by the four language models. We observe that, although LLMs can occasionally be misled by environmental distractions such as dead ends (as shown in Figure~\ref{fig:ChatGPT} and Figure~\ref{fig:DeepSeek}), the overall temporal ordering of the subgoals remains largely correct across all LLMs models we tested, even in complex environments like \textit{PointMaze-Medium}. Figure~\ref{fig:STOLLM} reports the averaged success rates over 100 trials for each of the LLM-generated subgoal sequences. These results demonstrate that, even when the subgoal sequence only partially preserves the correct temporal order, our STO-RL framework still achieves competitive performance. Therefore, our method exhibits robust general effectiveness across different language models. With the continued advancement of LLM capabilities, we anticipate further improvements in the performance of STO-RL.

\section{Conclusion}

In this work, we propose STO-RL, a novel framework for offline reinforcement learning guided by LLM-generated subgoal temporal order. Leveraging the reasoning capabilities of large language models (LLMs), STO-RL automatically decomposes language-based task instructions into temporally ordered subgoal sequences. Building upon this structured knowledge, we introduce a potential-based reward shaping scheme that incorporates subgoal temporal order to provide dense and informative feedback for offline policy training. This design effectively connects high-level task understanding with low-level policy learning, enabling agents to utilize subgoal progress as a shaping signal without requiring additional online interactions. Extensive experiments on discrete and continuous benchmarks (\textit{FourRoom, CliffWalking, PointMaze-UMaze}, and \textit{PointMaze-Medium}) show that STO-RL consistently accelerates learning, improves success rates, and yields shorter, more direct trajectories than state-of-the-art offline goal-conditioned and hierarchical RL baselines. Ablation studies further demonstrate its robustness to imperfect LLM-generated subgoal sequences.


Despite these promising results, STO-RL has certain limitations. First, it does not include mechanisms to refine or correct partially incorrect LLM-generated subgoals, which can restrict optimal performance in more complex environments. Second, the policies trained via STO-RL are task-specific and do not generalize to multiple tasks or domains. For future work, we will explore adaptive subgoal refinement strategies to improve temporal consistency, as well as extensions to multi-task and transfer learning settings. 




\begin{acks}
This work was partly supported by Natural Sciences and Engineering Research Council of Canada (NSERC). 
\end{acks}



\bibliographystyle{ACM-Reference-Format} 
\bibliography{sample}

\newpage
\section{Appendix}

\subsection{Proof of Lemma 1}
\label{sec::ProofL1}

\paragraph{Lemma 1.} \textit{(Equivalence of Returns for Equal-Length Successful Trajectories) Consider a goal-reaching task with $K$ temporally ordered subgoals. Any two successful trajectories with same episode length $T_L$ ($T_L\leq T$) yield equal discounted return $\sum_{t=0}^{T_L-1}\gamma^tr'(s_t,a_t,s_{t+1})$ under $r'(s_t,a_t,s_{t+1}) = r(s_t,a_t,s_{t+1}) + \gamma \Phi(s_{t+1}) - \Phi(s_t)$, where $\Phi(s_t) = -\frac{t}{T} \cdot \frac{1}{k_t}$.}

\begin{proof}
By \textit{Definition 1}, for trajectory $\tau$ with length $T_L$  we have $k_0 = 1, k_{T_L} = K$. The discounted return is $\sum_{t=0}^{T_L-1}\gamma^tr'(s_t,a_t,s_{t+1})$.

Since $r'(s_t,a_t,s_{t+1}) = r(s_t,a_t,s_{t+1}) + \gamma \Phi(s_{t+1}) - \Phi(s_t)$, we have:
\begin{equation}
    \begin{split}
        &\sum_{t=0}^{T_L-1}\gamma^tr'(s_t,a_t,s_{t+1}) = \sum_{t=0}^{T_L-1}\gamma^t (r(s_t,a_t,s_{t+1})+\gamma \Phi(s_{t+1}) - \Phi(s_t)) \\
        &=\sum_{t=0}^{T_L-1}\gamma^tr(s_t, a_t, s_{t+1})+\sum_{t=1}^{T_L}\gamma^t\Phi(s_{t}) - \sum_{t=0}^{T_L-1}\gamma^t\Phi(s_{t}) \\
        &=\sum_{t=0}^{T_L-1}\gamma^tr(s_t, a_t, s_{t+1}) + \gamma^{T_L}\Phi(s_{T_L}) - \Phi(s_{0}).
    \end{split}
\end{equation}

As $\Phi(s_t) = -\frac{t}{T} \cdot \frac{1}{k_t}$, $\Phi(s_{T_L}) = -\frac{T_L}{T} \cdot \frac{1}{K}$, $\Phi(s_{0}) =0$. Hence, $\sum_{t=0}^{T_L-1}\gamma^tr'(s_t,a_t,s_{t+1})$ is only related to episode length $T_L$.
\end{proof}

\subsection{Prompts to LLM for 4 Tasks in Experiments}
\label{sec::appPrompt}

In this section, we demonstrate the prompts to LLMs to generate temporally ordered subgoal sequences for all 4 tasks (\textit{CliffWalking, FourRoom, PointMaze-UMaze, PointMaze-Medium}) in our experiments. The default LLM model is ChatGPT 5.0. 
\subsubsection{CliffWalking}
\quad \\
Prompt $\Sigma$ to LLM:
\begin{tcolorbox}
You need to establish an ordered sub-task sequence for a CliffWalking Task, crossing a gridworld from Start State to Goal State while avoiding falling off a cliff. The map of maze is listed below: \\

The environment is a the 4x12 grid world.\\
The game starts with the player at location [3, 0].\\
The goal located at [3, 11]. \\
A cliff runs along [3, 1..10].\\
Your response should be like:\\
$\{$ SubTask 1: ‘Move to place’, containing states: “(1, 1), (1, 2),……”\\
,……,\\
‘SubTask N: ‘Move to goal’, containing states:”……”\\
$\}$\\
(Hint: the subtask sequence should cover all states in the maze map EXCEPT the walls)\\
(Hint: Each state can only be assigned to one sub-task)\\

\end{tcolorbox}
The LLM responses with:
\begin{tcolorbox}
$\{$
 SubTask 1: 'Navigate safely away from the start', \\
    containing states: "(3,0), (2,0)",\\
 SubTask 2: 'Move horizontally toward the goal while avoiding the cliff', \\
    containing states: "(0, 0), (0,1), (0,2), (0,3), (0,4), (0,5), (0,6), (0,7), (0,8), (0,9), (0,10), (0,11), (1,11), (1,10), (1,9), (1,8), (1,7), (1,6), (1,5), (1,4), (1,3), (1,2), (1,1), (1, 0), (2, 0), (2,1), (2,2), (2,3), (2,4),(2,5), (2,6), (2,7), (2,8), (2,9), (2,10) ",\\
 SubTask 3: 'Descend toward the goal once past the cliff', \\
    containing states: "(2, 11)",\\
 SubTask 4: 'Reach the goal precisely without stepping into the cliff', \\
    containing states: "(3,11)"\\
$\}$
\end{tcolorbox}

\subsubsection{PointMaze-UMaze}
\quad \\
Prompt $\Sigma$ to LLM:

\begin{tcolorbox}
You need to establish an ordered sub-task sequence for a Maze Navigation Task from Start State to Goal State. The map of maze is listed below:

\[
\text{U\_MAZE} = 
\begin{bmatrix}
1 & 1 & 1 & 1 & 1 \\
1 & r & 0 & 0 & 1 \\
1 & 1 & 1 & 0 & 1 \\
1 & g & 0 & 0 & 1 \\
1 & 1 & 1 & 1 & 1
\end{bmatrix}
\]

Where ‘r’ is the Start State, ‘g’ is the Goal State, ‘1’ are walls and ‘0’ are paths where the agent can move.\\

Your response should be like:\\
$\{$ SubTask 1: ‘Move to place’, containing states: “(1, 1), (1, 2),……”\\
,……,\\
‘SubTask N: ‘Move to goal’, containing states:”……”\\
$\}$\\

(Hint: the subtask sequence should cover all states in the maze map EXCEPT the walls)\\
(Hint: Each state can only be assigned to one sub-task)\\
\end{tcolorbox}

The LLM responses with:
\begin{tcolorbox}
    $\{$
SubTask 1: "Move along top corridor", containing states: "(1,1), (1,2)",\\
SubTask 2: "Move down the vertical connector", containing states: " (1,3), (2,3), (3,3)",\\
SubTask 3: "Move to goal (left along bottom)", containing states: "(3,2), (3,1)
$\}$
\end{tcolorbox}

\newpage

\subsubsection{FourRoom}
\quad \\
Prompt $\Sigma$ to LLM:
\begin{tcolorbox}
You need to establish an ordered sub-task sequence for a FourRoom Task to navigate from Start State to Goal State. The map of FourRoom is listed below:

\begingroup
\small
\setlength{\arraycolsep}{3pt}
\[
\text{FOUR\_ROOM} = 
\left[
\begin{array}{ccccccccccc}
r & 0 & 0 & 0 & 0 & 1 & 0 & 0 & 0 & 0 & 0 \\
0 & 0 & 0 & 0 & 0 & 1 & 0 & 0 & 0 & 0 & 0 \\
0 & 0 & 0 & 0 & 0 & 0 & 0 & 0 & 0 & 0 & 0 \\
0 & 0 & 0 & 0 & 0 & 1 & 0 & 0 & 0 & 0 & 0 \\
0 & 0 & 0 & 0 & 0 & 1 & 0 & 0 & 0 & 0 & 0 \\
1 & 1 & 0 & 1 & 1 & 1 & 1 & 1 & 0 & 1 & 1 \\
0 & 0 & 0 & 0 & 0 & 1 & 0 & 0 & 0 & 0 & 0 \\
0 & 0 & 0 & 0 & 0 & 1 & 0 & 0 & 0 & 0 & 0 \\
0 & 0 & 0 & 0 & 0 & 0 & 0 & 0 & 0 & 0 & 0 \\
0 & 0 & 0 & 0 & 0 & 1 & 0 & 0 & 0 & 0 & 0 \\
0 & 0 & 0 & 0 & 0 & 1 & 0 & 0 & 0 & 0 & g
\end{array}
\right]
\]
\endgroup

Where ‘r’ is the Start State, ‘g’ is the Goal State, ‘1’ are walls and ‘0’ are paths where the agent can move.\\

Your response should be like:\\
$\{$ SubTask 1: ‘Move to place’, containing states: “(1, 1), (1, 2),……”\\
,……,\\
‘SubTask N: ‘Move to goal’, containing states:”……”\\
$\}$\\

(Hint: the subtask sequence should cover all states in the maze map EXCEPT the walls)\\
(Hint: Each state can only be assigned to one sub-task)\\
\end{tcolorbox}

The LLM responses with:
\begin{tcolorbox}
    $\{$
 SubTask 1: 'Navigate Top-Left Room (Start Region)',\\
 containing states: [\\
 (0,0), (0,1), (0,2), (0,3), (0,4),\\
 (1,0), (1,1), (1,2), (1,3), (1,4),\\
 (2,0), (2,1), (2,2), (2,3), (2,4),\\
 (3,0), (3,1), (3,2), (3,3), (3,4),\\
 (4,0), (4,1), (4,2), (4,3), (4,4)\\
 ],

 SubTask 2: 'Traverse Middle Corridor and Two Corner Rooms (Top-Right + Bottom-Left)',\\
 containing states: [\\
 \# corridor and openings\\
 (2,5), (8,5), (5,2), (5,8),\\
 \# top-right room\\
 (0,6), (0,7), (0,8), (0,9), (0,10),\\
 (1,6), (1,7), (1,8), (1,9), (1,10),\\
 (2,6), (2,7), (2,8), (2,9), (2,10),\\
 (3,6), (3,7), (3,8), (3,9), (3,10),\\
 (4,6), (4,7), (4,8), (4,9), (4,10),\\
 \# bottom-left room\\
 (6,0), (6,1), (6,2), (6,3), (6,4),\\
 (7,0), (7,1), (7,2), (7,3), (7,4),\\
 (8,0), (8,1), (8,2), (8,3), (8,4),\\
 (9,0), (9,1), (9,2), (9,3), (9,4),\\
 (10,0), (10,1), (10,2), (10,3), (10,4)\\
 ],

 SubTask 3: 'Move in Bottom-Right Room (Goal Region)',\\
 containing states: [\\
 (6,6), (6,7), (6,8), (6,9), (6,10),\\
 (7,6), (7,7), (7,8), (7,9), (7,10),\\
 (8,6), (8,7), (8,8), (8,9), (8,10),\\
 (9,6), (9,7), (9,8), (9,9), (9,10),\\
 (10,6), (10,7), (10,8), (10,9), (10,10)\\
 ]
$\}$
\end{tcolorbox}

\newpage
\subsubsection{PointMaze-Medium}
\quad \\
Prompt $\Sigma$ to LLM:

\begin{tcolorbox}

You need to establish an ordered sub-task sequence for a Maze Navigation Task from Start State to Goal State. The map of maze is listed below:

\[
\text{MEDIUM\_MAZE} = \left[
\begin{array}{cccccccc}
1 & 1 & 1 & 1 & 1 & 1 & 1 & 1 \\
1 & r & 0 & 1 & 1 & 0 & 0 & 1 \\
1 & 0 & 0 & 1 & 0 & 0 & 0 & 1 \\
1 & 1 & 0 & 0 & 0 & 1 & 1 & 1 \\
1 & 0 & 0 & 1 & 0 & 0 & 0 & 1 \\
1 & 0 & 1 & 0 & 0 & 1 & 0 & 1 \\
1 & 0 & 0 & 0 & 1 & g & 0 & 1 \\
1 & 1 & 1 & 1 & 1 & 1 & 1 & 1 \\
\end{array}
\right]
\]
            
Where ‘r’ is the Start State, ‘g’ is the Goal State, ‘1’ are walls and ‘0’ are paths where the agent can move.\\

Your response should be like:\\
$\{$ SubTask 1: ‘Move to place’, containing states: “(1, 1), (1, 2),……”\\
,……,\\
‘SubTask N: ‘Move to goal’, containing states:”……”\\
$\}$\\

(Hint: the subtask sequence should cover all states in the maze map EXCEPT the walls)\\
(Hint: Each state can only be assigned to one sub-task)\\

\end{tcolorbox}
The LLM responses with:

\begin{tcolorbox}
$\{$
SubTask 1: 'Move from start and sweep left column', containing states: "(1,1), (1,2), (2,2), (2,1), (4,1), (5,1), (6,1), (6,2), (6,3)",\\
SubTask 2: 'Move to central cluster and clear center', containing states: "(5,3), (5,4), (4,4), (3,4), (3,3), (3,2), (4,2)",\\
SubTask 3: 'Move to upper-right corridor', containing states: "(2,4), (2,5), (2,6), (1,6), (1,5)",\\
SubTask 4: 'Sweep right-bottom and finish at goal', containing states: "(4,5), (4,6), (5,6), (6,6), (6,5)"
$\}$
\end{tcolorbox}

\subsubsection{Subgoal Sequences from Different LLM Models}
\quad \\
In the ablation study presented in Section 4.3, we evaluate STO-RL’s robustness to the reliability of LLM-generated subgoal sequences in the \textit{PointMaze-Medium} environment by using temporally ordered subgoals produced by different LLM models. Among these, several sequences contain minor inaccuracies, with one or two subgoals that do not contribute to achieving the final goal. This section demonstrates LLM's responses with such partially failed cases, highlighting the incorrect subgoals in red and explaining why they hinder successful goal reaching.

\paragraph{Example 1:}
\quad \\
    
\begin{tcolorbox}
$\{$ SubTask 1: ‘Move to left corridor’, \\
containing states: “(1, 1), (1, 2), (2, 1), (2, 2), (3, 2), (4, 1), (4, 2), (5, 1), (6, 1), (6, 2), (6, 3)”, \\
SubTask 2: ‘Move to central area’, containing states: “(3, 3), (3, 4), (4, 4), (5, 3), (5, 4)”
, \\
\textcolor{red}{SubTask 3: ‘Move to upper right area’, \\
containing states: “(1, 5), (1, 6), (2, 4), (2, 5), (2, 6)”,\\
(\textbf{This subgoal guide the agent to a dead end in PointMaze-Medium.})}\\
SubTask 4: ‘Move to goal’, \\
containing states: “(4, 5), (4, 6), (5, 6), (6, 5), (6, 6)”
    $\}$ 
\end{tcolorbox}

\paragraph{Example 2:}
\quad \\
\begin{tcolorbox}
    $\{$ 
SubTask 1: ‘Move to start area junction’, containing states: “(1,1), (1,2), (2,1), (2,2)”,\\
\textcolor{red}{SubTask 2: ‘Move to upper right corridor’, containing states: “(1,5), (1,6), (2,4), (2,5), (2,6)”,\\
(\textbf{This subgoal guides the agent into a dead end in \textit{PointMaze-Medium} environment and is not directly adjacent to Subtask 1.})\\}
SubTask 3: ‘Move to central corridor \& lower left area’, containing states: “(3,2), (3,3), (3,4), (4,1), (4,2), (5,1), (6,1), (6,2), (6,3), (5,3), (5,4)”,\\
SubTask 4: ‘Move to lower right corridor’, containing states: “(4,4), (4,5), (4,6), (5,6), (6,6)”,\\
SubTask 5: ‘Move to goal’, containing states: “(6,5)”
 $\}$ 
\end{tcolorbox}

Responses from LLMs contains both the ordered subgoal sequences and reliable mapping functions $h$ (i.e. the containing states in subtask descriptions) that maps the states to subgoal progress indices.

\subsection{Details of Environment Settings}

\paragraph{CliffWalking} CliffWalking is a classical goal-reaching task in \textit{Gymnasium}, represented as a $4\times12$ gridworld with a finite, discrete state space with 48 states and a finite discrete action space with 4 actions (move up, down, left, right). The agent starts at position $[3,0]$ and must reach the goal at $[3,11]$, while a cliff spanning $[3,1..10]$  sends the agent back to the start if entered. Episodes terminate upon reaching the goal. In our experiments, we fix the state and goal spaces, set a maximum episode length $T=100$, and modify the reward function to give a reward of 1 only when the goal state $[3,11]$ is reached, removing the default cliff penalty. For all methods (IQL, HIQL, GC-BC, and STO-RL), the state, action, and subgoal are one-hot encoded as inputs to the Q-network, value network, and policy network during training process. The offline dataset is generated using an expert policy trained via value iteration and a random policy, each executed with 50\% probability at every step, resulting in 1,000 trajectories with an average success rate of 0.50 and an average trajectory length of $69.94\pm32.82$.

\paragraph{FourRoom} FourRoom is a classical goal-reaching task in \textit{MiniGrid}, where the agent navigates a 2D maze composed of four interconnected rooms via four gaps in the walls. In our experiments, we construct an $11\times11$ gridworld, with each room sized $5\times5$ and walls at row and column 5. The gaps connecting the rooms are located at (5,2), (5,8), (2,5), and (8,5). The agent starts at $[0,0]$ and aims to reach the goal state at $[10,10]$, with a maximum episode length of $T=100$. The environment has a discrete state space of 104 states and a discrete action space of 4 actions (up, down, left, right). For all baselines (IQL, HIQL, GC-BC) and our STO-RL, the state, action, and subgoal are one-hot encoded as inputs to the Q-network, value network, and policy network during training. The offline dataset is generated using a combination of an expert policy trained via value iteration and a random policy, each executed with 50\% probability per step, yielding 1,000 trajectories with an average success rate of 0.137 and an average trajectory length of $97.21\pm9.22$.

\paragraph{PointMaze-UMaze} PointMaze-UMaze is a goal-reaching task in D4RL, where a 2-DoF ball, force-actuated along the $x$ and $y$ directions, must navigate a closed maze to reach a target goal. The maze is represented as a matrix encoding discrete cell positions $(i,j)$ with a side length of 1 m. Each cell can take one of five values: '0' for free paths, '1' for walls, 'r' for the start state, and 'g' for the goal state. The action space is continuous, defined as \textit{Box(-1.0, 1.0, (2,), float32)} in gymnasium, representing the linear force applied to the ball in the $x$ and $y$ directions. The state space is a 4-dimensional vector containing the ball’s position $(x,y)$ and corresponding linear velocities, with no constraints. In our experiments, we use a sparse reward, giving 1 when the ball reaches the target (considered reached if the Euclidean distance is less than 0.5 m) and set the maximum episode length as $T=200$. We employ the default map in \textit{PointMaze-UMaze-v3} with fixed start and goal states:
\[
\text{U\_MAZE} = 
\begin{bmatrix}
1 & 1 & 1 & 1 & 1 \\
1 & r & 0 & 0 & 1 \\
1 & 1 & 1 & 0 & 1 \\
1 & g & 0 & 0 & 1 \\
1 & 1 & 1 & 1 & 1
\end{bmatrix}
\]
where the start state is randomly initialized within the cell marked 'r' (range $[-1,1]$) with added noise, and the goal is similarly initialized within the cell marked 'g' (range $[-1,1]$) with noise. The noise is sampled from a 2D Gaussian distribution with standard deviations $(0.25,0.25)$. 

For the GC-BC baseline, subgoals are one-hot encoded as part of the input to the goal-conditioned policy network. To generate demonstration trajectories, we use a combination of an pre-trained expert policy via proximal policy optimization (PPO) and a random policy, with the random policy executed at 70\% probability and expert 30\% per step, resulting in 1,000 trajectories with an average success rate of 0.485 and an average trajectory length of $191.24\pm12.72$.

\paragraph{PointMaze-Medium}
PointMaze-Medium is a goal-reaching task in D4RL with larger maze maps, indicating more complexity. Similar to \textit{UMaze}, the agent is a 2-DoF ball, force-actuated along the $x$ and $y$ directions, which must navigate a closed maze to reach a target goal. The maze is represented as a matrix encoding discrete cell positions $(i,j)$ with a side length of 1 m. Each cell can take one of five values: '0' for free paths, '1' for walls, 'r' for the start state, and 'g' for the goal state. The action space is continuous, defined as \textit{Box(-1.0, 1.0, (2,), float32)} in gymnasium, representing the linear force applied to the ball in the $x$ and $y$ directions. The state space is a 4-dimensional vector containing the ball’s position $(x,y)$ and corresponding linear velocities, with no constraints. In our experiments, we use a sparse reward, giving 1 when the ball reaches the target (considered reached if the Euclidean distance is less than 0.5 m) and set the maximum episode length as $T=500$. We employ the default map in \textit{PointMaze-Medium-v3} with fixed start and goal states:
\[
\text{MEDIUM\_MAZE} = \left[
\begin{array}{cccccccc}
1 & 1 & 1 & 1 & 1 & 1 & 1 & 1 \\
1 & r & 0 & 1 & 1 & 0 & 0 & 1 \\
1 & 0 & 0 & 1 & 0 & 0 & 0 & 1 \\
1 & 1 & 0 & 0 & 0 & 1 & 1 & 1 \\
1 & 0 & 0 & 1 & 0 & 0 & 0 & 1 \\
1 & 0 & 1 & 0 & 0 & 1 & 0 & 1 \\
1 & 0 & 0 & 0 & 1 & g & 0 & 1 \\
1 & 1 & 1 & 1 & 1 & 1 & 1 & 1 \\
\end{array}
\right]
\]
            
where the start state is randomly initialized within the cell marked 'r' (range $[-2.5,2.5]$) with added noise, and the goal is similarly initialized within the cell marked 'g' (range $[1.5,-2.5]$) with noise. The noise is sampled from a 2D Gaussian distribution with standard deviations $(0.25,0.25)$. 

For the GC-BC baseline, subgoals are one-hot encoded as part of the input to the goal-conditioned policy network. To generate demonstration trajectories, we use a combination of an pre-trained expert policy via proximal policy optimization (PPO) and a random policy, with the random policy executed at 70\% probability and expert 30\% per step, resulting in 1,000 trajectories with an average success rate of 0.505 and an average trajectory length of $415.05\pm93.40$.

\end{document}